\documentclass[accepted]{uai2023} 

\usepackage[american]{babel}

\usepackage{natbib} 
\usepackage{mathtools} 
\usepackage{booktabs} 
\usepackage{tikz} 
\usepackage{graphicx}
\usepackage{amsmath}
\usepackage{amssymb}
\usepackage{booktabs}
\usepackage{multirow}

\usepackage[capitalize]{cleveref}
\crefname{section}{Sec.}{Secs.}
\Crefname{section}{Section}{Sections}
\Crefname{table}{Table}{Tables}
\crefname{table}{Tab.}{Tabs.}

\def\MCS{\mathop{\rm MCS}\nolimits}
\def\cwMCS{\mathop{\rm cwMCS}\nolimits}

\title{Two Sides of Miscalibration: Identifying Over and Under-Confidence Prediction for Network Calibration}

\author[1]{\href{mailto:<shuang.ao@open.ac.uk>?Subject=Your UAI 2023 paper}{Shuang Ao}{}}
\author[2]{Stefan Rueger}
\author[3]{Advaith Siddharthan}
\affil[1,2,3]{%
    Knowledge Media Institute.\\
    The Open University\\
    Milton Keynes, UK
}
  
\begin{document}
\maketitle
\begin{abstract}

Proper confidence calibration of deep neural networks is essential for reliable predictions in safety-critical tasks. Miscalibration can lead to model over-confidence and/or under-confidence; i.e., the model's confidence in its prediction can be greater or less than the model's accuracy. Recent studies have highlighted the over-confidence issue by introducing calibration techniques and demonstrated success on various tasks. However, miscalibration through under-confidence has not yet to receive much attention. In this paper, we address the necessity of paying attention to the under-confidence issue. We first introduce a novel metric, a miscalibration score, to identify the overall and class-wise calibration status, including being over or under-confident. Our proposed metric reveals the pitfalls of existing calibration techniques, where they often overly calibrate the model and worsen under-confident predictions. Then we utilize the class-wise miscalibration score as a proxy to design a calibration technique that can tackle both over and under-confidence. We report extensive experiments that show our proposed methods substantially outperforming existing calibration techniques. We also validate our proposed calibration technique on an automatic failure detection task with a risk-coverage curve, reporting that our methods improve failure detection as well as trustworthiness of the model. The code are available at \url{https://github.com/AoShuang92/miscalibration_TS}. 

\end{abstract}

\section{Introduction}
\label{sec:intro}

Effective confidence calibration for deep learning models improves the reliability of model predictions, which is imperative for practical deployment in safety-critical tasks including decision-making in medical diagnosis~\citep{raghu2019direct,dusenberry2020analyzing}, autonomous driving systems~\citep{o2020dependable, zhang2018deeproad}, assistance systems for socially vulnerable groups~\citep{ashmore2021assuring}, and nuclear power plant monitoring~\citep{linda2009neural}. 
A well-calibrated model implies that the confidence scores for its predictions reflect its accuracy over a dataset, so that the model does not exhibit high confidence with low accuracy and vice versa. Due to the capacity of the model and complexity of the dataset, the model inherits a certain degree of miscalibration during training. Intuitively, the average predicted confidence can be higher or lower than accuracy, namely over-confidence and under-confidence. It is essential to address both for reducing calibration errors. Prior studies focus on over-confidence alone as many deep neural networks have been shown to be over-confident in fields such as computer vision, natural language processing and object detection~\citep{muller2019does,guo2017calibration,lin2017focal}. Under-confidence, the other aspect of miscalibration, has not been addressed much in prior work. Under-confidence can be as problematic as over-confidence for applications such as automatic failure detection~\citep{hendrycks2017baseline, corbiere2019addressing, band2021benchmarking}. Failure detection is a primary assessment for safety-critical tasks, whereby the system can decide to keep its predictions, reject them, or hand them to human experts for verification. One popular way of setting the threshold is by model confidence~\citep{band2021benchmarking, corbiere2019addressing}, where data points that fall below the confidence threshold are considered unreliable predictions. If the model is under-confident, more samples will be predicted with lower confidence, including more correct predictions, unnecessarily increasing the  burden on human experts to verify, and thus limiting the practical applications of the decision-making system.

As interest in tackling miscalibration has increased, several calibration techniques have been introduced and proven to be effective, such as label smoothing~\citep{muller2019does}, temperature scaling~\citep{guo2017calibration}, focal loss~\citep{lin2017focal}, and Deep Ensembles~\citep{lakshminarayanan2017simple}. To categorize their methodologies, these techniques aim to soften or squeeze either ground truth, logits or predicted probabilities. Despite their success, these techniques mainly target over-confidence. For example, in temperature scaling,~\cite{guo2017calibration} pointed out that the temperature value $T$ should be greater than 1 to soften the softmax predictions. This statement might not be suitable when the model is under-confident, as the way to calibrate the under-confident model should be to scale up the confidence instead of softening them. Furthermore, a technique designed for over-confidence may worsen under-confident predictions, which could diminish the interpretability of the model.

There are several metrics introduced to quantify calibration errors of the predicted confidence, such as Expected Calibration Error (ECE)~\citep{naeini2015obtaining} and its variant to evaluate class-wise ECE~\citep{obadinma2021class}, Static Calibration Error (SCE)~\citep{nixon2019measuring}, Thresholded Adaptive Calibration Error (TACE)~\citep{nixon2019measuring} and Brier Score (BS)~\citep{ashukha2020pitfalls}. However, current evaluation metrics are unable to reflect the under or over-confidence status of the model, either in an overall or class-wise way, as they quantify the absolute calibration error. Identifying a model being over or under-confident in general, as well as class-wise miscalibration status, could better inform methofs to reduce calibration error.

In this work, we quantify models as being under and over-confident, both overall and class-wise by designing two novel metrics: Miscalibration Score (MCS) and class-wise Miscalibration Score (cwMCS). We use the class-wise Miscalibration score as the guidance to design calibration techniques that can tackle both over and under-confidence issues. In addition, we investigate pitfalls of the SOTA calibration techniques and find that most of them overly calibrate the model, leading to more severe under-confidence issues. Our contributions and findings are summarized as below: 

\begin{enumerate}

\item We design metrics to identify over and under-confidence, both overall and class-wise.
\item We propose novel calibration techniques by utilizing Miscalibration  score that can tackle both under and over-confidence issues.
\item With extensive experiments and observations we find that: (i) Unexpectedly, many of the recent SOTA models tend to be under-confident (e.g, Vision Transformers (ViT)~\citep{dosovitskiy2020image} and its variants); (ii) Many of existing calibration techniques can overly calibrate the model, leading to worse under-confidence; (iii) Our proposed calibration methods can benefit automatic failure detection by producing well-calibrated confidence estimates.

\end{enumerate}

\section{Related Work}
\label{sec:work}
\subsection{Calibration Techniques}
To alleviate the Miscalibration problem for deep neural networks, several techniques have been proposed and then widely applied as in-training or post-processing approaches. Label Smoothing (LS)~\citep{szegedy2016rethinking} reduces over-confidence by computing the cross-entropy loss with uniformly squeezed labels instead of one-hot labels. Extensive work of LS such as Margin-based Label Smoothing (MBLS)~\citep{liu2022devil} further provides a unifying constrained-optimization perspective of calibration losses. Focal Loss (FL)~\citep{lin2017focal} adds a focusing factor to the standard cross-entropy loss to deal with an imbalanced dataset. The recent work sample-dependent focal loss (FLSD)~\citep{mukhoti2020calibrating} investigate the effect of the loss on the training data and achieves impressive performance in calibration. Temperature Scaling (TS)~\citep{guo2017calibration} is a simple and straightforward method by applying a single-parameter variant of Platt Scaling~\citep{guo2017calibration} to logits produced from classifiers. For multi-class classification, a recent work~\citep{wenger2020non} employs a non-parametric representation with a latent Gaussian process; and Dirichlet distributions are utilized to calibrate probabilities from classifiers~\citep{kull2019beyond}.

\subsection{Calibration Metrics}
To measure the performance of calibration methods, the Expected Calibration Error (ECE)~\citep{naeini2015obtaining} was proposed and widely applied in various tasks, such as image classification~\citep{guo2017calibration,nixon2019measuring} and sentiment analysis~\citep{muller2019does,obadinma2021class}. ECE splits the data into bins, compares for each bin the average confidence and average accuracy, and averages over all bins.
Despite the success of ECE, studies have shown certain limitations of it. First, ECE is sensitive to the number of bins. Theoretically, the larger number of bins indicates a smaller bias in the approximation, but the variance will increase with the growing number of bins due to fewer data points to calculate in some bins~\citep{nixon2019measuring}. In addition, ECE is unable to reveal the class-wise ECE status as it runs on the data as a whole. More precisely, if under and over-confident data points fall into the same bin, the calibration error for this bin can be severely underestimated~\citep{nixon2019measuring}. To better assess the degree of model calibration,~\cite{obadinma2021class} proposed Contraharmonic Expected Calibration Error (CECE), Macro Subset Expected Calibration Error (MSECE) and Weighted Subset Expected Calibration Error (wsECE), respectively, to take the class-wise calibration error into consideration, especially for imbalanced datasets. However, the metric to reveal the miscalibration status as over or under-confident is still lacking in current calibration measurements.  

\subsection{Confidence Calibration for Failure Detection} One main practical goal for confidence calibration is failure detection, which allows the model to detect misclassified samples by filtering out samples with the low-confidence score. More specifically, failure detection aims at distinguishing correct and incorrect predictions based on their confidence or uncertainty ranking. With a fixed or learned pre-defined threshold, the erroneous samples will be rejected by the model, then passed to human experts or other systems for a second opinion. In other words, models with the reject option can abstain from making a prediction when they are likely to make a mistake~\citep{hendrickx2021machine}. It is thus beneficial to improve model calibration and increase a user’s trust in the system.
Failure detection has been widely used in medical and manufacturing fields, such as AI for breast cancer screening~\citep{leibig2022combining} and decision-making models for low-power Internet of Things (IoT) devices~\citep{cho2020leveraging}. Hence calibrated confidence is essential to verify the trustworthiness of the model. Despite the significance, how calibration could benefit misclassification detection still needs more attention.~\cite{zhu2022rethinking} claimed that most calibration techniques are harmful for failure detection as they worsen the confidence separation between correct and incorrect samples. In this work, we argue that with proper techniques, calibration can benefit failure detection in certain tasks.

\section{Methodology}
\label{sec:method}
\subsection{Preliminaries}
Consider input samples $X=\left\{x_1, x_2, \ldots, x_N\right\}$ with $N$ as the sample size and $K$ as the numbers of classes.

\paragraph{Temperature Scaling (TS)} As a simple extension of Platt scaling~\citep{guo2017calibration}, Temperature Scaling (TS)~\citep{guo2017calibration} is a post-hoc approach to rescale logits with a single scalar $T$. A logit vector $z$ is generated by a neural network with a given input $x$. $T$ is applied on the logit $z$ then it passes to the softmax function (denoted as $\sigma$), the new predicted confidence $\hat{p}^{TS}$ is:

\begin{equation}
\label{eq: ptstemp}
\hat{p}^{TS} = \max \: \sigma ( z / T)
\end{equation}

\paragraph{Expected Calibration Error (ECE)}
The Expected Calibration Error (ECE)~\citep{naeini2015obtaining} is one of our base evaluation metrics for calibration performance. This metric represents the absolute difference between the predicted model confidence and accuracy. The predictions are sorted and grouped into $M$ equal-spaced bins. Let $B_m$ denote the set of samples with predicted value falling into the $m^{th}$ bin, where the interval is $\left[\frac{i-1}{M}, \frac{i}{M}\right]$. Given $\mathbb{I}$ as the indicator function, the average accuracy $acc_m$ of $B_m$ is: $acc_m=\frac{1}{\left|B_m\right|} \sum_{i \in B_m} \mathbb{I}\left(\hat{\mathbf{y}}_i=\mathbf{y}_i\right)$. Likewise, the average confidence $conf_m$ of $B_m$ is calculated by averaging the confidence of all samples in the bin, which is written as $conf_m=\frac{1}{\left|B_m\right|} \sum_{i \in B_m} \hat{p}_i$. ECE then can be estimated by taking a weighted average of the absolute difference between the accuracy and confidence in each bin, which can be formulated as:


\begin{equation}
\label{equ: ece}
    ECE=\sum_m^M \frac{\left|B_m\right|}{N}\left|conf_m - acc_m\right|
\end{equation}

where $N$ is the number of samples. 

Note that ECE fails to show whether the the calibration error is caused by under or over-confident predictions. 

\paragraph{Weighted Subset Expected Calibration Error (wsECE)} 
To present the class-wise ECE, the data points are divided into bins in a class-wise way. First, divide the dataset $X,Y$ into $K$ subsets where each sample $x_n$ has the target $y = k$. Then pass each of these subsets to the ECE formula and get the result of individual ECEs with the size of $K$. Hence the class-wise ECE (cwECE) for each class is  the set:

\begin{equation}
\label{equ: cwece}
cwECE = \left\{ECE_1, \ ECE_2, \ \ldots, \ ECE_K\right\} 
\end{equation}

To represent the class-wise ECE comprehensively,~\cite{obadinma2021class} proposed the method called weighted subset Expected Calibration Error (wsECE). As in imbalanced datasets, calibration error is unlikely to be the same for all classes. Error on minority classes can be much worse than the overall calibration performance. Hence the weight of each subset is significant to measure the overall calibration. Each subset ECEs is weighted by the percentage of data points of the class, and wsECE calculates the weighted average of all ECEs. Suppose the sample size in each class is $\left\{n_1, n_2, \ldots, n_K\right\}$, the calculation of wsECE can be written as:

\begin{equation}
\label{equ:wsECE}
wsECE=\frac{n_1}{N} ECE_1+\frac{n_2}{N} ECE_2+\ldots \frac{n_K}{N} ECE_K
\end{equation}

wsECE gives an insight into the calibration performance of each class, including marginal classes of most under and over-confident ones. However, wsECE also does not distinguish between under and over-confidence. 

\subsection{Proposed Methods}

In this section, we design a novel metric, Miscalibration Score (MCS), to measure the calibration error, taking into account whether the model is under or over-confident. We also extend MCS to weighted subset MCS by considering the sample size in each class for imbalanced datasets. 
Then we integrate MCS with Temperature Scaling (TS) to address limitations with traditional TS techniques. 

\paragraph{Miscalibration Score (MCS)} To reveal the under and over confidence levels in each class, we first take the difference $conf_m - acc_m$ instead of the absolute value shown in equation~\eqref{equ: ece}. In this case, positive values indicate over-confidence, and negative values indicate under-confidence. With this slight modification, the Miscalibration Score (MCS) is formulated as:

\begin{equation}
\label{equ: msc}
    MCS=\sum_m^M \frac{\left|B_m\right|}{N}\left(conf_m-acc_m\right)
\end{equation}

To obtain the class-wise MCS (cwMCS), similar to the calculated method of cwECE (Eq. \ref{equ: cwece}), the MCS for each subset is $cwMCS = \left\{MCS_1, MCS_2, \ldots, MCS_K\right\}$, which clearly shows the level of over or under confidence for each class. 

The MCS can be positive (+) for over-confidence and negative (-) for under-confidence. Given, a dataset with $K$ classes in total, the number of under-confident classes is $k^-$, the number of over-confident classes is $k^+$, the sample size for each under-confident class is $\left\{n_{1^-}, n_{2^-}, \ldots, n_{k^-}\right\}$, and the sample size for each over-confident class is $\left\{n_{1^+}, n_{2^+}, \ldots, n_{k^+}\right\}$. The corresponding MCS in each under and over-confidence class are:
$\left\{MCS_{1^-}, MCS_{2^-}, \ldots, MCS_{k^-}\right\}$ and $\left\{MCS_{1^+}, MCS_{2^+}, \ldots, MCS_{k^+}\right\}$.

Due to the learning dynamics of the model and the complexity of the dataset, even for the balanced dataset, the number of under and over-confident classes is not likely to be equal. Therefore, inspired by the wsECE~\citep{obadinma2021class}, which considers the class size, we introduce a weighted subset MCS (wsMCS) to address severe imbalanced Miscalibration status (under or over-confidence) in predictions. First, we group all under-confidence and over-confidence classes and calculate their MCS with the sample size in each class, namely $wsMCS^-$ and $wsMCS^+$. The miscalibration error from the group with minority classes can be worse than what is indicated by the overall calibration, which can dominate the entire calibration performance. Then, to avoid this issue, we take the under and over-confidence class number into consideration to obtain overall wsMCS, and the final formula can be written as:


\begin{align}
\label{equ: wsmsc}
wsMCS^+ &=\frac{n_{1^+}}{N} \MCS_{1^+}+ \ldots + \frac{n_{k^+}}{N} \MCS_{k^+}\\
wsMCS^- &=\frac{n_{1^-}}{N} \MCS_{1^-}+ \ldots + \frac{n_{k^-}}{N} \MCS_{k^-}\\
wsMCS &= \frac{k^+}{K} wsMCS^+ + \frac{k^-}{K} wsMCS^-
\end{align}

\paragraph{cwMCS Temperature Scaling (cwMCS TS)} A scalar value of temperature ($T$) is unable to fulfill the calibration requirement for both over and under-confident classes, especially for marginal cases. We hereby integrate the class-wise Miscalibration Score (cwMCS) with $T$ to formulate a vector ($T^{\cwMCS}$) with the size of $K$:

\begin{equation}
\label{eq:newT}
    T^{\cwMCS} = T \cdot (1+\gamma \cdot \cwMCS)
\end{equation}

where $cwMCS$ is max-normalized, $\gamma$ is the weighting parameter to control the effect of class-wise miscalibration score into temperature $T$, which is tuned in the range of $>-1$ to $<+1$ with the increment of 0.001. 
The new confidence prediction $\hat{p}^{T^{\cwMCS}}$ with $T^{\cwMCS}$ is now:

\begin{equation}
\label{eq: ptcwmcstemp}
\hat{p}^{T^{\cwMCS}} = \max \: \sigma\left( z / T^{\cwMCS}\right)
\end{equation}

where $\sigma$ is the softmax function. As $MCS$ can be negative or positive, this allows for scaling up or scaling down, which is capable to fix under-confidence as well as over-confidence. Specifically, if a class is under-confident and the MCS is negative, the cwMCS TS ($T^{\cwMCS}$) will be smaller than the T and vice versa. It is theoretically possible for MCS to be 0. Under the class-wise context, if the MCS becomes 0, it means no impact on the $T$ of that very class. In other words, $T$ will be used to recalibrate that class without modifications in our $T^{\cwMCS}$ for that class. When it comes to weighted subset MCS (wsMCS), only the class with MCS being 0 has no impact on the positive or negative MCS calculation.

\begin{table*}[!h]
\caption{Results of baselines, TS and our proposed class-wise miscalibration score TS (cwMCS TS). The image datasets IN, Tiny-IN, C100 and C10 refer to ImageNet, Tiny-ImageNet, CIFAR100 and CIFAR10, and text dataset SST2, EMO, SENT mean Stanford Sentiment Treebank v2, TWEETEVAL Emotion and Sentiment respectively. ECE and wsECE are shown as percentages. wsECE is weighted subset ECE, and wsMCS is our weighted subset miscalibration score. Best calibration results for each row are shown in bold. The best result for wsMCS is calculated using the absolute value.}
\centering
\label{main_results}
\scalebox{.75}{
\begin{tabular}{c|c|c|ccc|ccc|ccc}
\hline
Dataset                   & Model        & Acc (\%) & \multicolumn{3}{c|}{Baseline}                                               & \multicolumn{3}{c|}{TS}                                                               & \multicolumn{3}{c}{cwMCS TS}                                                               \\ \hline
                          &              &          & \multicolumn{1}{c|}{ECE(\%)} & \multicolumn{1}{c|}{wsECE(\%)}      & wsMCS  & \multicolumn{1}{c|}{ECE(\%)}       & \multicolumn{1}{c|}{wsECE(\%)} & wsMCS           & \multicolumn{1}{c|}{ECE(\%)}       & \multicolumn{1}{c|}{wsECE(\%)}      & wsMCS           \\ \hline
\multirow{13}{*}{\rotatebox[origin=c]{90}{IN}}      & ViT          & 83.30    & \multicolumn{1}{c|}{2.20}    & \multicolumn{1}{c|}{13.40}          & 0.393  & \multicolumn{1}{c|}{0.90}          & \multicolumn{1}{c|}{14.00}     & -0.353          & \multicolumn{1}{c|}{\textbf{0.70}} & \multicolumn{1}{c|}{13.80}          & \textbf{-0.219} \\ \cline{2-12} 
                          & SwinT        & 84.40    & \multicolumn{1}{c|}{8.70}    & \multicolumn{1}{c|}{18.50}          & -3.354 & \multicolumn{1}{c|}{2.10}          & \multicolumn{1}{c|}{14.10}     & -0.384          & \multicolumn{1}{c|}{\textbf{1.40}} & \multicolumn{1}{c|}{\textbf{13.80}} & \textbf{-0.135} \\ \cline{2-12} 
                          & DeiT         & 81.30    & \multicolumn{1}{c|}{6.40}    & \multicolumn{1}{c|}{18.60}          & -2.477 & \multicolumn{1}{c|}{3.30}          & \multicolumn{1}{c|}{15.30}     & 1.620           & \multicolumn{1}{c|}{\textbf{2.90}} & \multicolumn{1}{c|}{\textbf{15.30}} & \textbf{0.180}  \\ \cline{2-12} 
                          & CaiT         & 83.00    & \multicolumn{1}{c|}{3.60}    & \multicolumn{1}{c|}{15.80}          & -0.815 & \multicolumn{1}{c|}{3.10}          & \multicolumn{1}{c|}{14.90}     & -0.032          & \multicolumn{1}{c|}{\textbf{2.90}} & \multicolumn{1}{c|}{\textbf{14.90}} & \textbf{-0.002} \\ \cline{2-12} 
                          & BeiT         & 85.90    & \multicolumn{1}{c|}{5.70}    & \multicolumn{1}{c|}{16.10}          & -2.292 & \multicolumn{1}{c|}{4.20}          & \multicolumn{1}{c|}{13.40}     & -0.503          & \multicolumn{1}{c|}{\textbf{2.00}} & \multicolumn{1}{c|}{\textbf{13.20}} & \textbf{-0.335} \\ \cline{2-12} 
                          & CoaT         & 80.70    & \multicolumn{1}{c|}{7.60}    & \multicolumn{1}{c|}{19.50}          & -2.916 & \multicolumn{1}{c|}{3.00}          & \multicolumn{1}{c|}{15.80}     & 0.940           & \multicolumn{1}{c|}{\textbf{2.20}} & \multicolumn{1}{c|}{\textbf{15.70}} & \textbf{0.126}  \\ \cline{2-12} 
                          & CrossViT     & 81.90    & \multicolumn{1}{c|}{5.20}    & \multicolumn{1}{c|}{17.90}          & -2.060 & \multicolumn{1}{c|}{2.80}          & \multicolumn{1}{c|}{15.50}     & 0.950           & \multicolumn{1}{c|}{\textbf{2.50}} & \multicolumn{1}{c|}{\textbf{15.50}} & \textbf{0.071}  \\ \cline{2-12} 
                          & ConvMixer      & 79.30    & \multicolumn{1}{c|}{17.40}   & \multicolumn{1}{c|}{25.10}          & -6.720 & \multicolumn{1}{c|}{8.90}          & \multicolumn{1}{c|}{16.20}     & -0.780          & \multicolumn{1}{c|}{\textbf{0.80}} & \multicolumn{1}{c|}{\textbf{16.10}} & \textbf{-0.254} \\ \cline{2-12} 
                          & ConvNext     & 83.50    & \multicolumn{1}{c|}{3.60}    & \multicolumn{1}{c|}{15.00}          & -0.294 & \multicolumn{1}{c|}{3.40}          & \multicolumn{1}{c|}{14.60}     & 0.174           & \multicolumn{1}{c|}{\textbf{1.30}} & \multicolumn{1}{c|}{\textbf{4.20}}  & \textbf{0.078}  \\ \cline{2-12} 
                          & ResNet34     & 71.20    & \multicolumn{1}{c|}{4.00}    & \multicolumn{1}{c|}{\textbf{18.30}} & 1.116  & \multicolumn{1}{c|}{\textbf{1.60}} & \multicolumn{1}{c|}{18.40}     & \textbf{-0.004} & \multicolumn{1}{c|}{\textbf{1.60}} & \multicolumn{1}{c|}{18.40}          & \textbf{0.105}  \\ \cline{2-12} 
                          & DenseNet121  & 71.80    & \multicolumn{1}{c|}{3.00}    & \multicolumn{1}{c|}{\textbf{18.20}} & 0.182  & \multicolumn{1}{c|}{1.40}          & \multicolumn{1}{c|}{18.40}     & \textbf{-0.017} & \multicolumn{1}{c|}{\textbf{1.30}} & \multicolumn{1}{c|}{18.40}          & -0.097          \\ \cline{2-12} 
                          & VGG16        & 69.90    & \multicolumn{1}{c|}{3.30}    & \multicolumn{1}{c|}{18.30}          & 0.908  & \multicolumn{1}{c|}{\textbf{1.80}} & \multicolumn{1}{c|}{18.50}     & \textbf{-0.046} & \multicolumn{1}{c|}{\textbf{1.80}} & \multicolumn{1}{c|}{18.50}          & -0.049          \\ \cline{2-12} 
                          & EfficientNet & 75.60    & \multicolumn{1}{c|}{14.50}   & \multicolumn{1}{c|}{24.80}          & -5.508 & \multicolumn{1}{c|}{14.40}         & \multicolumn{1}{c|}{25.60}     & -4.409          & \multicolumn{1}{c|}{\textbf{1.40}} & \multicolumn{1}{c|}{\textbf{20.70}} & \textbf{0.018}  \\ \hline
\multirow{3}{*}{\rotatebox[origin=c]{90}{Tiny-IN}} & ResNet34     & 59.70    & \multicolumn{1}{c|}{9.30}    & \multicolumn{1}{c|}{18.30}          & 0.330  & \multicolumn{1}{c|}{8.30}          & \multicolumn{1}{c|}{16.90}     & -0.140          & \multicolumn{1}{c|}{\textbf{4.30}} & \multicolumn{1}{c|}{\textbf{14.90}} & \textbf{-0.034} \\ \cline{2-12} 
                          & DenseNet121  & 61.80    & \multicolumn{1}{c|}{13.00}   & \multicolumn{1}{c|}{19.10}          & 0.507  & \multicolumn{1}{c|}{9.20}          & \multicolumn{1}{c|}{16.40}     & -0.330          & \multicolumn{1}{c|}{\textbf{2.10}} & \multicolumn{1}{c|}{\textbf{14.40}} & \textbf{-0.032} \\ \cline{2-12} 
                          & VGG16        & 61.60    & \multicolumn{1}{c|}{15.60}   & \multicolumn{1}{c|}{20.10}          & 0.617  & \multicolumn{1}{c|}{8.29}         & \multicolumn{1}{c|}{16.60}     & -0.032          & \multicolumn{1}{c|}{\textbf{2.80}} & \multicolumn{1}{c|}{\textbf{10.20}} & \textbf{-0.024} \\ \hline
\multirow{3}{*}{\rotatebox[origin=c]{90}{C100}}   & ResNet34     & 70.80    & \multicolumn{1}{c|}{13.40}   & \multicolumn{1}{c|}{16.10}          & 0.134  & \multicolumn{1}{c|}{5.60}          & \multicolumn{1}{c|}{11.60}     & 0.062           & \multicolumn{1}{c|}{\textbf{1.60}} & \multicolumn{1}{c|}{\textbf{1.20}}  & \textbf{-0.010} \\ \cline{2-12} 
                          & DenseNet121  & 66.20    & \multicolumn{1}{c|}{14.80}   & \multicolumn{1}{c|}{17.70}          & 0.147  & \multicolumn{1}{c|}{7.20}          & \multicolumn{1}{c|}{12.30}     & -0.060          & \multicolumn{1}{c|}{\textbf{2.10}} & \multicolumn{1}{c|}{\textbf{4.60}}  & \textbf{-0.010} \\ \cline{2-12} 
                          & VGG16        & 64.30    & \multicolumn{1}{c|}{9.90}    & \multicolumn{1}{c|}{14.00}          & 0.097  & \multicolumn{1}{c|}{5.40}          & \multicolumn{1}{c|}{11.10}     & -0.076          & \multicolumn{1}{c|}{\textbf{1.30}} & \multicolumn{1}{c|}{\textbf{1.10}}  & \textbf{-0.006} \\ \hline
\multirow{3}{*}{\rotatebox[origin=c]{90}{C10}}     & ResNet34     & 90.20    & \multicolumn{1}{c|}{4.10}    & \multicolumn{1}{c|}{4.60}           & 0.041  & \multicolumn{1}{c|}{1.10}          & \multicolumn{1}{c|}{2.60}      & 0.002           & \multicolumn{1}{c|}{\textbf{0.50}} & \multicolumn{1}{c|}{\textbf{1.50}}  & \textbf{0.001}  \\ \cline{2-12} 
                          & DenseNet121  & 81.70    & \multicolumn{1}{c|}{11.30}   & \multicolumn{1}{c|}{11.30}          & 0.114  & \multicolumn{1}{c|}{1.70}          & \multicolumn{1}{c|}{4.30}      & 0.012           & \multicolumn{1}{c|}{\textbf{0.80}} & \multicolumn{1}{c|}{\textbf{2.20}}  & \textbf{0.001}  \\ \cline{2-12} 
                          & VGG16        & 87.60    & \multicolumn{1}{c|}{7.10}    & \multicolumn{1}{c|}{7.40}           & 0.071  & \multicolumn{1}{c|}{3.70}          & \multicolumn{1}{c|}{3.50}      & 0.037           & \multicolumn{1}{c|}{\textbf{1.70}} & \multicolumn{1}{c|}{\textbf{1.60}}  & \textbf{0.012}  \\ \hline
\multirow{3}{*}{\rotatebox[origin=c]{90}{SST-2}}      & RoBERTa      & 94.00    & \multicolumn{1}{c|}{4.29}    & \multicolumn{1}{c|}{4.60}           & 0.081  & \multicolumn{1}{c|}{3.21}          & \multicolumn{1}{c|}{2.80}      & 0.045           & \multicolumn{1}{c|}{\textbf{1.21}} & \multicolumn{1}{c|}{\textbf{1.70}}  & \textbf{0.005}  \\ \cline{2-12} 
                          & BERT         & 92.40    & \multicolumn{1}{c|}{6.09}    & \multicolumn{1}{c|}{6.20}           & 0.107  & \multicolumn{1}{c|}{3.10}          & \multicolumn{1}{c|}{4.60}      & 0.064           & \multicolumn{1}{c|}{\textbf{1.09}} & \multicolumn{1}{c|}{\textbf{2.60}}  & \textbf{0.014}  \\ \cline{2-12} 
                          & DistilBERT   & 91.10    & \multicolumn{1}{c|}{8.00}    & \multicolumn{1}{c|}{8.00}           & 0.147  & \multicolumn{1}{c|}{4.17}          & \multicolumn{1}{c|}{6.90}      & 0.069           & \multicolumn{1}{c|}{\textbf{3.56}} & \multicolumn{1}{c|}{\textbf{5.00}}  & \textbf{0.054}  \\ \hline
\multirow{3}{*}{\rotatebox[origin=c]{90}{EMO}}   & RoBERTa      & 83.40    & \multicolumn{1}{c|}{3.66}    & \multicolumn{1}{c|}{7.20}           & 0.194  & \multicolumn{1}{c|}{2.70}          & \multicolumn{1}{c|}{6.20}      & \textbf{0.036}  & \multicolumn{1}{c|}{\textbf{1.84}} & \multicolumn{1}{c|}{\textbf{5.80}}  & 0.074           \\ \cline{2-12} 
                          & BERT         & 81.00    & \multicolumn{1}{c|}{4.52}    & \multicolumn{1}{c|}{8.40}           & -0.096 & \multicolumn{1}{c|}{1.66}          & \multicolumn{1}{c|}{7.50}      & 0.070           & \multicolumn{1}{c|}{\textbf{1.29}} & \multicolumn{1}{c|}{\textbf{7.20}}  & \textbf{0.058}  \\ \cline{2-12} 
                          & DistilBERT   & 80.20    & \multicolumn{1}{c|}{15.70}   & \multicolumn{1}{c|}{19.50}          & 0.765  & \multicolumn{1}{c|}{6.52}          & \multicolumn{1}{c|}{9.80}      & 0.303           & \multicolumn{1}{c|}{\textbf{4.31}} & \multicolumn{1}{c|}{\textbf{8.20}}  & \textbf{0.260}  \\ \hline
\multirow{2}{*}{\rotatebox[origin=c]{90}{SENT}}  & RoBERTa      & 72.40    & \multicolumn{1}{c|}{2.16}    & \multicolumn{1}{c|}{\textbf{7.90}}  & 0.048  & \multicolumn{1}{c|}{1.96}          & \multicolumn{1}{c|}{8.00}      & \textbf{-0.018} & \multicolumn{1}{c|}{\textbf{1.54}} & \multicolumn{1}{c|}{8.00}           & 0.039           \\ \cline{2-12} 
                          & xlm\_RoBERTa & 68.20    & \multicolumn{1}{c|}{4.64}    & \multicolumn{1}{c|}{9.60}           & 0.052  & \multicolumn{1}{c|}{3.35}          & \multicolumn{1}{c|}{10.80}     & -0.084          & \multicolumn{1}{c|}{\textbf{2.10}} & \multicolumn{1}{c|}{\textbf{9.00}}  & \textbf{-0.023} \\ \hline
\end{tabular}}
\end{table*}

\section{Experimental Setup}

\subsection{Datasets and Baselines}
We validate the proposed calibration method with four benchmark image classification datasets: ImageNet 2012 (IN)~\citep{russakovsky2015imagenet}, Tiny-ImageNet (Tiny-IN)~\citep{deng2009imagenet}, and CIFAR10 (C10) and CIFAR100 (C100)~\citep{krizhevsky2009learning}. For baselines, we use state-of-the-art (SOTA) Vision Transformer (ViT)~\citep{dosovitskiy2020image} and its variants such as Swin-Transformer (SwinT)~\citep{liu2021swin}, Data-efficient Image Transformers (DeiT)~\citep{touvron2021training}, Class-Attention in Image Transformers (CaiT)~\citep{touvron2021going},  Bidirectional Encoder representation from Image Transformers (BEiT)~\citep{bao2021beit}, Co-scale Conv-attentional image Transformers (CoaT)~\citep{xu2021co}, Cross-Attention Multi-Scale Vision Transformer (CrossViT)~\citep{chen2021crossvit}, ConvMixer~\citep{trockman2022patches} with the ImageNet pretrained weights from TIMM~\footnote{https://github.com/rwightman/pytorch-image-models} library. To report comprehensive results on various models architectures, we also use the Convolutional neural networks (CNNs) in our experiments, namely EfficientNet~\citep{tan2019efficientnet}, DenseNet121~\citep{huang2017densely}, ResNet34~\citep{he2016deep}, and VGG16~\citep{simonyan2014very}. All models are with pretrained weights of ImageNet dataset. For recent SOTA calibration techniques MBLS~\citep{liu2022devil} and FLSD~\citep{mukhoti2020calibrating}, we utilize the pre-trained model and official implementation from the repository~\footnote{https://github.com/by-liu/MbLS}. Furthermore, for the comparison of ECE results on additional SOTA calibation techniques Platt Scaling (Platt), Isotonic Regression (Isotonic)~\citep{zadrozny2002transforming}, Beta calibration (Beta)~\citep{kull2017beta}, Bayesian Binning into Quantile (BBQ)~\citep{naeini2015obtaining}, Temperature Scaling (TS) and Gaussian Process Calibration (GPCalib)~\citep{wenger2020non}, we utilize the pre-trained model and official implementation from the repository~\footnote{https://github.com/JonathanWenger/pycalib}. 

We also verify our proposed methods on three text classification dataset: Stanford Sentiment Treebank v2 (SST-2)~\citep{socher2013recursive}, TWEETEVAL-Emotion and TWEETEVAL-Sentiment~\citep{barbieri2020tweeteval}. For baselines, we use BERT~\citep{devlin2018bert}, RoBERTa~\citep{liu2019roberta} and DistilBERT~\citep{sanh2019distilbert} with Hugging Face implementations~\footnote{https://huggingface.co}.

As our experiment utilises a post-processing approach, and to avoid overfitting to obtain the temperature, we primarily utilize each dataset's test set. For the ImageNet dataset, we equally divide its original test set of 50,000 images into validation and test sets for a fair comparison. For Tiny-ImageNet, CIFAR10/100 and all three text dataset, an 80/10/10 for training/validation/test split is applied. The dataset setting is the same as other experiments for the investigation of existing calibration techniques (shown in table~\ref{tab:sota techniques}).

\subsection{Implementation Details}

For reproducability and fair comparison, we utilized existing publicly available pretrained weights where possible for our investigation and experimentation. For example, all the observations with ImageNet and existing calibration techniques are from pretrained weights except Tiny-ImageNet and CIFAR10/100 dataset are applied for model training and testing with the networks of ResNet34, DenseNet121 and VGG16. The models are trained with stochastic gradient descent (SGD) optimizer with momentum of 0.9 and weight decay of 0.0005. The initial learning rate for ResNet34 and DenseNet121 is 0.1, and 0.001 for VGG16. The batch size for all training is 1024 and for validation and test is 2048. The GPU of the Nvidia Tesla P40 is used for all experiments. For the evaluation metric of ECE, wsECE and our proposed method, we set the number of bins as $M=15$.  

\begin{figure*}[!h]
\centerline{\includegraphics[width=0.9\textwidth]{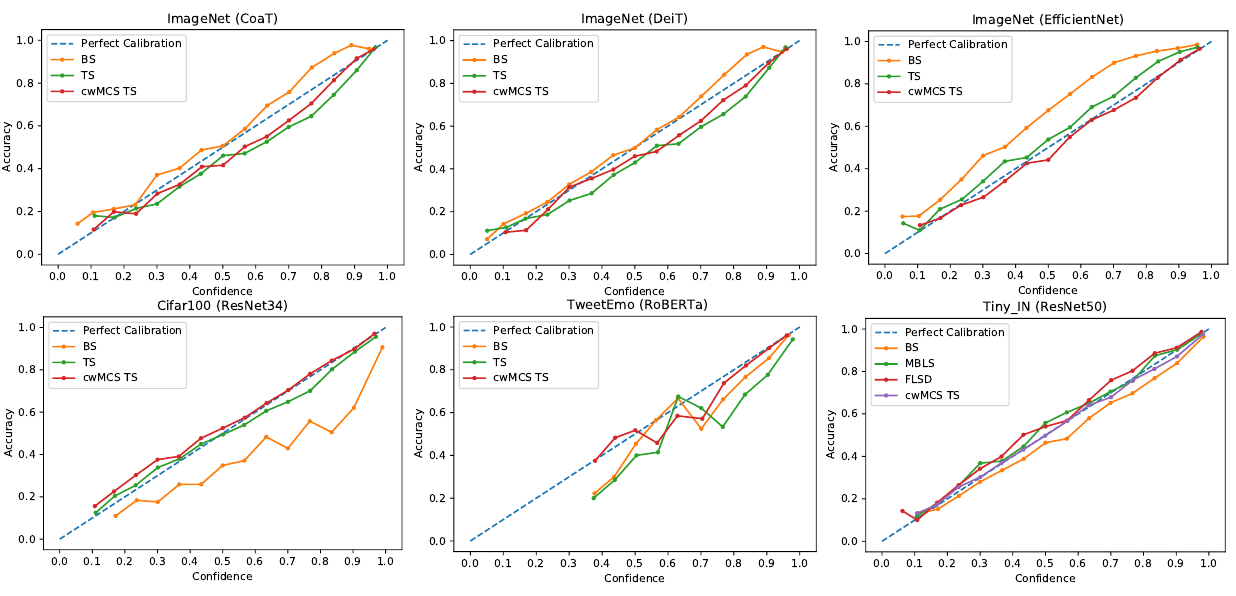}}
\caption{Reliability Diagram (RD) for ImageNet, Tiny-ImageNet, CIFAR100 and TweetEval-Emotion with calibration techniques. BS, TS, MBLS, FLSD, cwMCS TS mean baseline, temperature scaling, margin-based label smoothing, sample-dependent focal loss and class-wise Miscalibration Score TS respectively. First row from left to right: RD for ImageNet with CoaT, DeiT and EfficientNet. Second row from left to right: RD for CIFAR100 with ResNet34, TweetEval-Emotion with RoBERTa, and the comparison of SOTA calibration techniques on Tiny-ImageNet with RestNet50.}
\label{fig:rd}
\end{figure*}

\begin{table*}[!ht]
\caption{Results of average under-confidence Miscalibration score (UC MCS) and average over-confidence miscalibration score (OC MCS) for baseline, TS and our proposed method cwMCS TS.} 

\label{tab:u_o msc}
\centering
\scalebox{.75}{
\begin{tabular}{lc|cc|cc|cc}
\hline
\multicolumn{2}{l|}{}                                                                                                      & \multicolumn{2}{c|}{Baseline}                                                 & \multicolumn{2}{c|}{TS}                                                     & \multicolumn{2}{c}{cwMCS TS}                                                  \\ \hline
\multicolumn{1}{c|}{Dataset}                                                                  & \multicolumn{1}{c|}{Model} & \multicolumn{1}{l|}{UC MCS (\%)}         & \multicolumn{1}{l|}{OC MCS (\%)} & \multicolumn{1}{l|}{UC MCS (\%)}       & \multicolumn{1}{l|}{OC MCS (\%)} & \multicolumn{1}{l|}{UC MCS (\%)}         & \multicolumn{1}{l}{OC MCS (\%)} \\ \hline
\multicolumn{1}{l|}{\multirow{4}{*}{IN}}                                               & ViT                        & \multicolumn{1}{c|}{-4.2 (48.3)}          & 7.4 (51.7)                        & \multicolumn{1}{c|}{-5.2 (61.3)}        & 7.2 (38.7)                        & \multicolumn{1}{c|}{\textbf{-3.7 (58.3)}} & \textbf{0.5 (41.7)}              \\ \cline{2-8} 
\multicolumn{1}{l|}{}                                                                         & SwinT                       & \multicolumn{1}{c|}{-11.6 (85.8)}         & 9.2 (14.2)                        & \multicolumn{1}{c|}{-5.3 (64)}          & 9.2 (36)                          & \multicolumn{1}{c|}{\textbf{-0.5 (65.7)}} & \textbf{8.7 (34.3)}              \\ \cline{2-8} 
\multicolumn{1}{l|}{}                                                                         & DeiT                       & \multicolumn{1}{c|}{-10.4 (79.6)}         & 9.0 (20.4)                        & \multicolumn{1}{c|}{-5.2 (54.7)}        & 9.6 (45.3)                        & \multicolumn{1}{c|}{\textbf{-3.2 (55.3)}} & \textbf{5.6 (44.7)}              \\ \cline{2-8} 
\multicolumn{1}{l|}{}                                                                         & ConvMixer                    & \multicolumn{1}{c|}{-19.0 (90.4)}         & \textbf{8.1 (9.6)}                & \multicolumn{1}{c|}{-8.9 (61.5)}        & 8.6 (38.5)                        & \multicolumn{1}{c|}{\textbf{-5.7 (59.7)}} & 8.4 (40.3)                       \\ \hline
\multicolumn{1}{l|}{\multirow{3}{*}{\begin{tabular}[c]{@{}l@{}}Tiny-IN \end{tabular}}} & ResNet34                   & \multicolumn{1}{c|}{\textbf{-3.1 (11.5)}} & 10.6 (88.5)                       & \multicolumn{1}{c|}{-5.8 (58)}          & 6.3 (42)                          & \multicolumn{1}{c|}{-5.4 (52)}            & \textbf{4.3 (48)}                \\ \cline{2-8} 
\multicolumn{1}{l|}{}                                                                         & DenseNet121                & \multicolumn{1}{c|}{\textbf{-1.0 (2.5)}}           & 13.4 (97.5)                       & \multicolumn{1}{c|}{-5.5 (58.5)}        & 7.6 (41.5)                        & \multicolumn{1}{c|}{-5.2 (56.5)} & \textbf{6.6 (43.5)}              \\ \cline{2-8} 
\multicolumn{1}{l|}{}                                                                         & VGG16                      & \multicolumn{1}{c|}{\textbf{-0.8 (1.5)}}  & 15.9 (98.5)                       & \multicolumn{1}{c|}{-5.2 (57.5)}        & 5.1 (42.5)                        & \multicolumn{1}{c|}{-4.9 (52.5)}          & \textbf{3.1 (47.5)}              \\ \hline
\multicolumn{1}{l|}{\multirow{3}{*}{C100}}                                                & Res34                      & \multicolumn{1}{c|}{\textbf{-0.1 (5)}}    & 13.4 (95)                         & \multicolumn{1}{c|}{-5.2 (61)}          & 6.2 (39)                          & \multicolumn{1}{c|}{-5.0 (57)}            & \textbf{4.2 (43)}                \\ \cline{2-8} 
\multicolumn{1}{l|}{}                                                                         & DenseNet121                & \multicolumn{1}{c|}{\textbf{-1.3 (1)}}    & 15 (99)                           & \multicolumn{1}{c|}{-5.7 (55)}          & 6.3 (46)                          & \multicolumn{1}{c|}{-4.3 (58)}            & \textbf{5.6 (42)}                \\ \cline{2-8} 
\multicolumn{1}{l|}{}                                                                         & VGG16                      & \multicolumn{1}{c|}{\textbf{-1.3 (3)}}    & 10.3 (97)                         & \multicolumn{1}{c|}{-4.4 (56)}          & 8.7 (44)                          & \multicolumn{1}{c|}{-4.4 (56)}            & \textbf{3.7 (44)}                \\ \hline
\multicolumn{1}{l|}{\multirow{3}{*}{C10}}                                                 & ResNet34                   & \multicolumn{1}{c|}{0.0 (0)}              & 4.1 (100)                         & \multicolumn{1}{c|}{\textbf{-1.4 (30)}} & 2.1 (70)                          & \multicolumn{1}{c|}{-1.5 (30)}            & \textbf{1.5 (70)}                \\ \cline{2-8} 
\multicolumn{1}{l|}{}                                                                         & DenseNet121                & \multicolumn{1}{c|}{0.0 (0)}              & 11.3 (100)                        & \multicolumn{1}{c|}{-1.6 (40)}          & 5.4 (60)                          & \multicolumn{1}{c|}{\textbf{-0.2 (30)}}   & \textbf{5.2 (70)}                \\ \cline{2-8} 
\multicolumn{1}{l|}{}                                                                         & VGG16                      & \multicolumn{1}{c|}{0.0 (0)}              & 7.1 (100)                         & \multicolumn{1}{c|}{-0.9 (40)}          & 2.8 (60)                          & \multicolumn{1}{c|}{\textbf{-0.7 (50)}}   & \textbf{0.2 (50)}                \\ \hline
\end{tabular}}
\end{table*}

\section{Results}
\label{sec:results}
In this work, we first conduct an experiment to understand prevalence of under-confidence using our proposed metric of wsMCS. Then we observe the pitfalls of the existing calibration techniques and the limitations of the conventional calibration metric of ECE. Finally, we show that cwMCS based TS can effectively tackle both under and over-confidence. We also validate the efficacy of our cwMCS TS on a real-world application of automatic failure detection. All the experiment and results are shown in the Tables of~\ref{main_results}, ~\ref{tab:u_o msc} and~\ref{tab:sota techniques}, and Figures of~\ref{fig:rd} and~\ref{fig:curve}.

\subsection{Identifying Under-confidence}
To identify under-confident predictions, we report for several Transformers, CNNs and BERT-based architectures the accuracy, ECE, wsECE and proposed wsMCS. Table~\ref{main_results} presents the performance with image and text classification. Surprisingly, wsMCS for most of the Transformer and BERT-based architectures are negative for baselines which reveals their under-confidence in predictions. It is also observed that the Transformer-based architecture (e.g., ConvMixer) with lower accuracy yield higher ECE or wsECE but these metrics are unable to reveal the specific direction of the miscalibration. On the other hand, our wsMCS clearly indicates under-confidence. In terms of CNN-based architecture for image classification, the wsMCS results suggest that most of these predictions are over-confident. In Table~\ref{tab:u_o msc}, almost more than half of the classes become under-confident after applying TS, indicating that TS can overly calibrate models. Our proposed method cwMCS TS substantially improves the mean over-confidence score and contributes to better calibration for under-confident classes. Additional observations of under-confidence can be found in appendix A.

\begin{table}[!h]
\caption{The comparison of SOTA calibration techniques and ours. Datasets Tiny-IN, C10 refer to Tiny-ImageNet and CIFAR10. All models are with pretrained weights of ResNet50 from~\cite{liu2022devil}. UC and OC are average under and over-confidence miscalibration score. wsMCS is overall miscalibration score and Acc is accuracy. All results except wsMCS are shown in percentage for clarity.}
\label{tab:sota techniques}
\scalebox{.75}{
\begin{tabular}{c|c|c|c|c|c|c}
\hline
Dataset                   & Method & ECE(\%) & wsMCS  & UC(\%) & OC(\%) & Acc(\%) \\ \hline
\multirow{7}{*}{\rotatebox[origin=c]{90}{Tiny-IN}} & CE     & 3.77    & 6.356  & -4.00  & 6.60   & 64.80   \\ \cline{2-7} 
                          & TS     & 1.40    & 1.340  & -4.50  & 5.60   & 64.80   \\ \cline{2-7} 
                          & LS     & 3.07    & -5.094 & -60.00 & 4.80   & 65.50   \\ \cline{2-7} 
                          & MBLS   & 1.86    & -3.288 & -5.50  & 5.20   & 64.70   \\ \cline{2-7} 
                          & FL     & 3.10    & -5.365 & -6.30  & 4.30   & 63.20   \\ \cline{2-7} 
                          & FLSD   & 2.84    & -5.466 & -3.90  & 5.00   & 63.90   \\ \cline{2-7} 
                          & Ours   & 1.05    & 0.640  & -1.90  & 5.00   & 64.90   \\ \hline
\multirow{7}{*}{\rotatebox[origin=c]{90}{C10}}      & CE     & 6.51    & 0.650  & 0.00   & 65.00  & 92.40   \\ \cline{2-7} 
                          & TS     & 5.34    & 0.532  & 0.00   & 32.00  & 92.40   \\ \cline{2-7} 
                          & LS     & 2.79    & 0.332  & -8.40  & 4.10   & 94.90   \\ \cline{2-7} 
                          & MBLS   & 1.49    & 0.290  & -0.40  & 2.70   & 94.70   \\ \cline{2-7} 
                          & FL     & 3.78    & -0.339 & -4.20  & 0.30   & 94.30   \\ \cline{2-7} 
                          & FLSD   & 3.61    & -0.334 & -33.40 & 0.00   & 94.20   \\ \cline{2-7} 
                          & Ours   & 1.38    & 0.224  & -2.50  & 22.00  & 92.40   \\ \hline
\end{tabular}}
\end{table}

\subsection{Pitfalls of Existing Calibration Techniques}

We first investigate several widely applied calibration techniques of LS, FL, TS, and recent works of MBLS and FLSD to check their calibration performance. Table~\ref{tab:sota techniques} shows the ECE, wsMCS, average under-confidence and over-confidence MCS and accuracy for each calibration method. The cross-entropy (CE) baseline is always over-confident as the wsMCS is positive for all experiments. However, after applying calibration techniques, wsMCS of FL and FLSD turns to negative values in results both dataset, which indicates the model is overly calibrated. On the other hand, the traditional calibration metric, ECE, cannot reveal the issue as shown in Table~\ref{tab:sota techniques}.

We further investigate TS with the most recent architectures of ViT and its variants with IN, Tiny-IN, C10/100 datasets in Table~\ref{main_results}. Similar to other calibration techniques, TS also tends to over-calibrate the Transformer-based models. We also observe that more than half the classes become under-confident on average after the calibration (as shown in Table~\ref{tab:u_o msc}). These results suggest that models are overly calibrated as existing calibration techniques are designed to focus on the over-confidence issue. This is also observed in the reliability diagram in Figure~\ref{fig:rd}. For example, after applying TS in ImageNet dataset with EfficientNet, FLSD and MBLS in Tiny-ImageNet dataset with ResNet50, the model turns into under-confidence as the curve of accuracy is clearly higher than the confidence. Therefore, techniques that can tackle both under and over-confident issues are required for better calibration. In addition, we conduct extensive experiments comparing our method with more calibration techniques in terms of ECE, which is shown in Table~\ref{addtional_results}. Our method yields consistently superior performance over almost all additional SOTA and recent techniques.


\subsection{Effectiveness of proposed calibration method}
Table~\ref{main_results} also shows the calibration results for TS and the proposed cwMCS TS validated with four image and three text dataset and various model architectures. In most experiments, our proposed methods outperform baselines and the TS method, with substantial improvement of ECE and Miscalibration score. Notably, the accuracy remains the same for all methods in each experiment.


\subsection{Application to failure detection}
To verify the effectiveness of our proposed method cwMCS TS on failure detection (FD), we utilize the predictive uncertainty to distinguish correct from incorrect samples by following~\citep{corbiere2019addressing}. More specifically, we calculate the entropy of the confidence for each sample. Higher entropy means lower confidence of the model prediction, and all data is ranked based on this method. A smaller entropy means a bigger distance between class predictions within a predicted distribution, which suggests a higher probability for correct predictions. In addition, entropy is a symmetric measure regardless of the position in a distribution, where [0.8, 0.2] and [0.2, 0.8] produce the same entropy. It may cause an issue for class probabilities as each probability points to a target. From our observation, our approach can successfully discriminate the incorrect from correct predictions regardless of similar distributions. 

\begin{figure*}[!h]
\centerline{\includegraphics[width=1\textwidth]{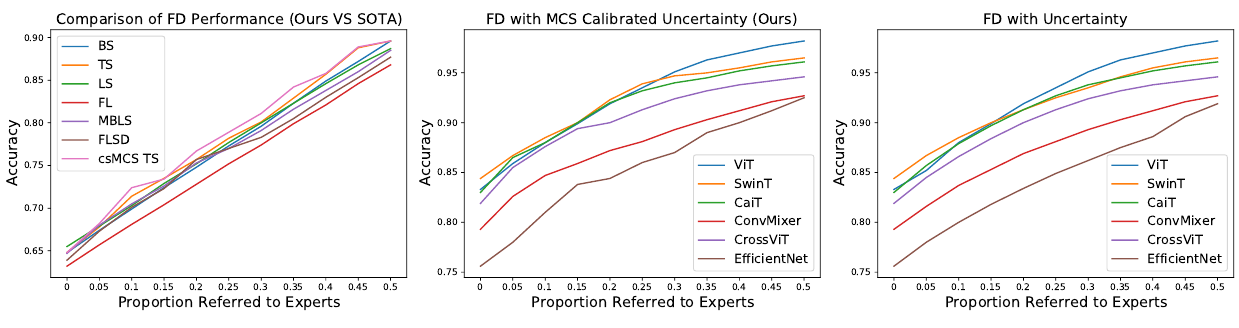}}

\caption{Risk-coverage Curves for failure detection (FD) performance. "Proportion Referred to Experts (x-axis)" means the percentage of data discarded, and "Accuracy (y-axis)" is the accuracy of remaining samples of the test set. The higher accuracy indicates better calibration of the model, as samples with lower confidence are removed properly. Left: the comparison of FD on SOTA calibration techniques and ours on ResNet50 with Tiny-ImageNet. Middle: FD for ImageNet with calibrated models (our method); Right: FD for ImageNet with uncalibrated models (baseline).}
\label{fig:curve}
\end{figure*}

\begin{table*}[!ht]
\caption{The ECE result on models with various SOTA calibration techniques on ImageNet dataset: with uncalibrated (Uncal.), Platt Scaling (Platt), Isotonic Regression (Isotonic), Beta calibration (Beta), Bayesian Binning into Quantile (BBQ), Temperature Scaling (TS), Gaussian Process Calibration (GPCalib) and our method. ECE reslut is shown in percentage for clarity.}
\label{addtional_results}
\centering
\scalebox{.85}{

\begin{tabular}{l|l|l|l|l|l|l|l|l}
\hline
Model      & {\color[HTML]{333333} Uncal.} & Platt & Isotonic & Beta  & BBQ   & TS   & GPCalib       & Ours          \\ \hline
DesNet121  & 3.00                          & 9.49  & 26.82    & 22.97 & 15.12 & 1.40 & 3.57          & \textbf{1.30} \\ \hline
VGG19      & 3.57                          & 9.65  & 28.10    & 24.16 & 16.17 & 1.94 & 3.64          & \textbf{1.79} \\ \hline
ResNet50   & 4.21                          & 8.75  & 27.24    & 22.50 & 16.35 & 3.43 & 3.35          & \textbf{1.95} \\ \hline
ResNeXt50  & 5.96                          & 8.89  & 26.84    & 17.89 & 19.90 & 4.84 & 2.79          & \textbf{2.60} \\ \hline
ResNeXt101 & 6.24                          & 8.53  & 28.44    & 16.31 & 14.96 & 4.23 & \textbf{2.50} & 2.88          \\ \hline
\end{tabular}}
\end{table*}

Figure~\ref{fig:curve} shows the risk-coverage curves for Tiny-ImageNet with ResNet50 with the application of SOTA calibration techniques (Left), and ImageNet dataset accessed with vanilla (Right) and calibrated (Middle) architectures. The different percentages of 'Referred to Experts' means the samples removed from the test set, and the accuracy is calculated with the remaining data. We use the predictive uncertainty to rank all the samples in the test set, where samples with high uncertainty will be referred. A higher accuracy rates is better, as it suggests errors are concentrated in the low-confidence samples; i.e the method is better at discriminating correct and incorrect samples. 

In Figure~\ref{fig:curve} (left), with the increased percentage referred to experts, only our proposed method cwMCS and TS beat the baseline, where our proposed method outperforms TS especially between 0.15 and 0.4 proportion. LS has the highest accuracy with the entire test set, but it quicky falls behind when just 5\% samples are removed. FLSD reaches the same performance as TS with the proportion of 0.2, then later falls behind LS and MBLS. The results shows that our proposed method performs better in error detection than SOTA calibration methods. 

Figure~\ref{fig:curve} (Right) shows the FD performance of uncalibrated models, where SwinT has the highest accuracy before any data is removed. However, SwinT is surpassed at the point of proportion 0.15, then ViT outperforms all other models with the increment of samples referred. After calibration with our proposed method (shown in Figure~\ref{fig:curve} (Middle)), the curves all rise more steeply until 0.15, and SwinT maintains the highest accuracy until proportion 0.3, indicating better trustworthiness with our method.




\section{Discussion}
\label{sec:discuss}
In this work, we introduced a novel Miscalibration Score (MCS) to quantify over or under-confident predictions of models, both overall and class-wise. One limitation of the overall MCS is that it mostly indicates the trend of over or under-confidence, but not the magnitude of the miscalibration error. The reason is that we average the actual values of the difference between prediction confidence and accuracy instead of the absolute values. This results in lower scores as negative and positive (under and over-confidence) cancel each other out. However, the model still needs to be calibrated. To fix this issue, we calculate MCS for over and under-confident classes separately, namely average under-confidence MCS and average over-confidence MCS  respectively, as shown in Table~\ref{tab:u_o msc}. The corresponding percentage of classes being under or over-confident is also presented in parentheses, from where we can tell how many classes contribute to each calibration status. However, this percentage distribution does not decide whether a model is well-calibrated or not. In other words, neither more under-confidence classes nor more over-confidence classes could guarantee less calibration error. The better calibration happens when their absolute value tends towards zero. 

After quantifying the over and under-confidence, overall and class-wise (cwMCS), we utilize the cwMCS to design our MCS-aware calibration technique. Our proposed method effectively improves both ECE and wsECE (shown in Table~\ref{main_results}), as well as tackles under and over-confidence issues separately (shown in Table~\ref{tab:u_o msc}). The reliability diagram in Figure~\ref{fig:rd} also validates this. 

Even though some studies have claimed that the recent state-of-the-art deep neural networks tend to be over-confident, our findings in Table~\ref{main_results} and ~\ref{tab:u_o msc} are quite contrary. In the baseline of ViT and its variants, the overall MCS shown in Table~\ref{main_results} are mostly negative, and Table~\ref{tab:u_o msc} suggests that the percentage of under-confident classes are much higher than over-confident ones. Furthermore, by comparing the recent calibration techniques and our proposed method, we find that many calibration methods overly calibrate the model. Table~\ref{tab:sota techniques} and Figure~\ref{fig:rd} both show that the model turns under-confident after applying these techniques. For automatic failure detection result in Figure~\ref{fig:curve}, the risk-coverage curve indicates that our method can filter incorrect samples more effectively within a smaller proportion. Hence solving calibration issue, especially the under-confidence, can improve the trustworthiness of the model. It also suggests that with proper calibration techniques, model confidence can help detect misclassification. 

\section{Conclusion}
\label{sec:conclusion}

In conclusion, our work highlights that under-confidence is as crucial as over-confidence and hence should receive more attention from researchers. We introduced a novel metric Miscalibration Score (MCS), to identify under and over-confidence, overall and class-wise. Then we design class-wise MCS-aware calibration techniques to tackle both under and over-confidence issues. With extensive experiments, we validate that our proposed calibration technique can effectively address different miscalibration statuses. In addition, we observe that many of the current calibration techniques overly calibrate the model, inclined to a more severe under-confidence issue. Furthermore, our proposed method substantially improves error detection and proves that proper confidence calibration techniques can benefit failure detection for safety-critical tasks. For future work, we will apply our MCS-aware method to more calibration techniques, such as label smoothing and focal loss. As our experiments focus on image and text classification, we will also apply our proposed method to other tasks, such as text generation and image segmentation.



\bibliography{ao_503}
\appendix

\appendix

\onecolumn
\begin{center}
    \huge{Supplementary Material}
\end{center}
\section*{Identifying Under-confidence}
\label{supp_oc_uc}

\begin{table*}[!ht]
\caption{Results of average under-confidence mis-calibration score (UC MCS) and average over-confidence miscalibration score (OC MCS) for baseline, TS and our proposed method cwMCS TS. All results are shown in percentage for clarity. Best results for each row are shown in bold. The value in the bracket shows the percentage of class being under or over-confident.} 

\label{tab:u_o msc}
\scalebox{.9}{
\begin{tabular}{lc|cc|cc|cc}
\hline
\multicolumn{2}{l|}{}                                                                                                      & \multicolumn{2}{c|}{Baseline}                                                 & \multicolumn{2}{c|}{TS}                                                     & \multicolumn{2}{c}{cwMCS TS}                                                  \\ \hline
\multicolumn{1}{c|}{Dataset}                                                                  & \multicolumn{1}{c|}{Model} & \multicolumn{1}{l|}{UC MCS (\%)}         & \multicolumn{1}{l|}{OC MCS (\%)} & \multicolumn{1}{l|}{UC MCS (\%)}       & \multicolumn{1}{l|}{OC MCS (\%)} & \multicolumn{1}{l|}{UC MCS (\%)}         & \multicolumn{1}{l}{OC MCS (\%)} \\ \hline
\multicolumn{1}{l|}{\multirow{13}{*}{IN}}                                               & ViT                        & \multicolumn{1}{c|}{-4.2 (48.3)}          & 7.4 (51.7)                        & \multicolumn{1}{c|}{-5.2 (61.3)}        & 7.2 (38.7)                        & \multicolumn{1}{c|}{\textbf{-3.7 (58.3)}} & \textbf{0.5 (41.7)}              \\ \cline{2-8} 
\multicolumn{1}{l|}{}                                                                         & SwinT                       & \multicolumn{1}{c|}{-11.6 (85.8)}         & 9.2 (14.2)                        & \multicolumn{1}{c|}{-5.3 (64)}          & 9.2 (36)                          & \multicolumn{1}{c|}{\textbf{-0.5 (65.7)}} & \textbf{8.7 (34.3)}              \\ \cline{2-8} 
\multicolumn{1}{l|}{}                                                                         & DeiT                       & \multicolumn{1}{c|}{-10.4 (79.6)}         & 9.0 (20.4)                        & \multicolumn{1}{c|}{-5.2 (54.7)}        & 9.6 (45.3)                        & \multicolumn{1}{c|}{\textbf{-3.2 (55.3)}} & \textbf{5.6 (44.7)}              \\ \cline{2-8} 
\multicolumn{1}{l|}{}                                                                         & CaiT                       & \multicolumn{1}{c|}{-6.7 (67.6)}          & 9.8 (32.4)                        & \multicolumn{1}{c|}{-5.2 (57.6)}        & 9.6 (42.4)                        & \multicolumn{1}{c|}{\textbf{-3.2 (58.2)}} & \textbf{7.6 (41.8)}              \\ \cline{2-8} 
\multicolumn{1}{l|}{}                                                                         & BeiT                       & \multicolumn{1}{c|}{-9.0 (81.9)}          & 8.9 (18.1)                        & \multicolumn{1}{c|}{-5.4 (65.3)}        & 8.5 (34.7)                        & \multicolumn{1}{c|}{\textbf{-4.1 (62.2)}} & \textbf{8.2 (37.8)}              \\ \cline{2-8} 
\multicolumn{1}{l|}{}                                                                         & CoaT                       & \multicolumn{1}{c|}{-10.9 (83.1)}         & 8.7 (16.9)                        & \multicolumn{1}{c|}{-6.8 (55.6)}        & 9.0 (44.4)                        & \multicolumn{1}{c|}{\textbf{-5.4 (57.8)}} & \textbf{8.0 (42.2)}              \\ \cline{2-8} 
\multicolumn{1}{l|}{}                                                                         & CrossViT                   & \multicolumn{1}{c|}{-9.6 (76.8)}          & 9.5 (23.2)                        & \multicolumn{1}{c|}{-5.5 (56.1)}        & 9.6 (43.9)                        & \multicolumn{1}{c|}{\textbf{-3.5 (56.1)}} & \textbf{8.6 (43.9)}              \\ \cline{2-8} 
\multicolumn{1}{l|}{}                                                                         & ConvMix                    & \multicolumn{1}{c|}{-19.0 (90.4)}         & \textbf{8.1 (9.6)}                & \multicolumn{1}{c|}{-8.9 (61.5)}        & 8.6 (38.5)                        & \multicolumn{1}{c|}{\textbf{-5.7 (59.7)}} & 8.4 (40.3)                       \\ \cline{2-8} 
\multicolumn{1}{l|}{}                                                                         & ConvNext                   & \multicolumn{1}{c|}{-5.9 (59.8)}          & 9.3 (40.2)                        & \multicolumn{1}{c|}{-5.3 (53.8)}        & 9.2 (46.2)                        & \multicolumn{1}{c|}{\textbf{-4.3 (51.6)}} & \textbf{9.0 (48.4)}              \\ \cline{2-8} 
\multicolumn{1}{l|}{}                                                                         & ResNet34                   & \multicolumn{1}{c|}{\textbf{-4.8 (40.1)}} & 9.9 (59.9)                        & \multicolumn{1}{c|}{-6.4 (53.8)}        & 8.6 (46.2)                        & \multicolumn{1}{c|}{-6.3 (52)}            & \textbf{8.5 (48)}                \\ \cline{2-8} 
\multicolumn{1}{l|}{}                                                                         & DenseNet121                & \multicolumn{1}{c|}{\textbf{-5.2 (43.7)}} & 9.3 (56.3)                        & \multicolumn{1}{c|}{-6.3 (55.4)}        & \textbf{8.4 (44.6)}               & \multicolumn{1}{c|}{-6.4 (55.2)}          & \textbf{8.4 (44.8)}              \\ \cline{2-8} 
\multicolumn{1}{l|}{}                                                                         & VGG16                      & \multicolumn{1}{c|}{\textbf{-5.3 (40.5)}} & 8.8 (59.5)                        & \multicolumn{1}{c|}{-6.3 (53.9)}        & 8.1 (46.1)                        & \multicolumn{1}{c|}{-6.2 (54)}            & \textbf{8.0 (46)}                \\ \cline{2-8} 
\multicolumn{1}{l|}{}                                                                         & EfficientNet               & \multicolumn{1}{c|}{-17.0 (90.3)}         & 8.4 (9.7)                         & \multicolumn{1}{c|}{-16.0 (87.4)}       & 8.2 (12.6)                        & \multicolumn{1}{c|}{\textbf{-8.8 (53.6)}} & \textbf{1.2 (46.4)}              \\ \hline
\multicolumn{1}{l|}{\multirow{3}{*}{\begin{tabular}[c]{@{}l@{}}Tiny-IN \end{tabular}}} & ResNet34                   & \multicolumn{1}{c|}{\textbf{-3.1 (11.5)}} & 10.6 (88.5)                       & \multicolumn{1}{c|}{-5.8 (58)}          & 6.3 (42)                          & \multicolumn{1}{c|}{-5.4 (52)}            & \textbf{4.3 (48)}                \\ \cline{2-8} 
\multicolumn{1}{l|}{}                                                                         & DenseNet121                & \multicolumn{1}{c|}{-1.0 (2.5)}           & 13.4 (97.5)                       & \multicolumn{1}{c|}{-5.5 (58.5)}        & 7.6 (41.5)                        & \multicolumn{1}{c|}{\textbf{-5.2 (56.5)}} & \textbf{6.6 (43.5)}              \\ \cline{2-8} 
\multicolumn{1}{l|}{}                                                                         & VGG16                      & \multicolumn{1}{c|}{\textbf{-0.8 (1.5)}}  & 15.9 (98.5)                       & \multicolumn{1}{c|}{-5.2 (57.5)}        & 5.1 (42.5)                        & \multicolumn{1}{c|}{-4.9 (52.5)}          & \textbf{3.1 (47.5)}              \\ \hline
\multicolumn{1}{l|}{\multirow{3}{*}{C100}}                                                & Res34                      & \multicolumn{1}{c|}{\textbf{-0.1 (5)}}    & 13.4 (95)                         & \multicolumn{1}{c|}{-5.2 (61)}          & 6.2 (39)                          & \multicolumn{1}{c|}{-5.0 (57)}            & \textbf{4.2 (43)}                \\ \cline{2-8} 
\multicolumn{1}{l|}{}                                                                         & DenseNet121                & \multicolumn{1}{c|}{\textbf{-1.3 (1)}}    & 15 (99)                           & \multicolumn{1}{c|}{-5.7 (55)}          & 6.3 (46)                          & \multicolumn{1}{c|}{-4.3 (58)}            & \textbf{5.6 (42)}                \\ \cline{2-8} 
\multicolumn{1}{l|}{}                                                                         & VGG16                      & \multicolumn{1}{c|}{\textbf{-1.3 (3)}}    & 10.3 (97)                         & \multicolumn{1}{c|}{-4.4 (56)}          & 8.7 (44)                          & \multicolumn{1}{c|}{-4.4 (56)}            & \textbf{3.7 (44)}                \\ \hline
\multicolumn{1}{l|}{\multirow{3}{*}{C10}}                                                 & ResNet34                   & \multicolumn{1}{c|}{0.0 (0)}              & 4.1 (100)                         & \multicolumn{1}{c|}{\textbf{-1.4 (30)}} & 2.1 (70)                          & \multicolumn{1}{c|}{-1.5 (30)}            & \textbf{1.5 (70)}                \\ \cline{2-8} 
\multicolumn{1}{l|}{}                                                                         & DenseNet121                & \multicolumn{1}{c|}{0.0 (0)}              & 11.3 (100)                        & \multicolumn{1}{c|}{-1.6 (40)}          & 5.4 (60)                          & \multicolumn{1}{c|}{\textbf{-0.2 (30)}}   & \textbf{5.2 (70)}                \\ \cline{2-8} 
\multicolumn{1}{l|}{}                                                                         & VGG16                      & \multicolumn{1}{c|}{0.0 (0)}              & 7.1 (100)                         & \multicolumn{1}{c|}{-0.9 (40)}          & 2.8 (60)                          & \multicolumn{1}{c|}{\textbf{-0.7 (50)}}   & \textbf{0.2 (50)}                \\ \hline
\end{tabular}}
\end{table*}

Table~\ref{tab:u_o msc} illustrates the results of mean under-confidence and mean over-confidence scores, as well as the percentage of classes with different confidence statuses correspondingly. For ImageNet dataset with transformers variants, most of the classes are under-confident with baselines, where the absolute value of mean under-confidence score is higher than the mean over-confidence score. The model with the highest percentage of under-confident classes is ConvMix, where only 10 percent of classes are over-confident. When it comes to CNNs, over and under-confident classes are more balanced. Surprisingly, EfficientNet has a similar behavior as Convmix, where the percentage of under-confident classes are much higher than over-confident ones. After applying TS, the percentage of over and under-confident classes are more balanced, and our proposed method cwMCS TS keeps this trend. Compared to baseline, our cwMCS TS method almost halves over and under-confidence scores, whereas TS only makes a slight change of them. For Tiny-ImageNet, CIFAR100 and CIFAR10 datasets with CNNs, more than 90 percent of classes are over-confident in baselines, with none of the classes under-confident for CIFAR10 dataset. However, more than half of the classes become under-confident after applying TS, indicating that TS can overly calibrate models. Our proposed method cwMCS TS significantly improves the mean over-confidence score and contributes to better calibration for under-confident classes.



\end{document}